\documentclass[a4paper]{article}
\usepackage{INTERSPEECH2020}
\usepackage{graphicx}
\usepackage{amsfonts}
\usepackage{amssymb}
\usepackage{amsmath}
\usepackage{multirow}
\usepackage{enumitem}

\title{Incremental Machine Speech Chain\\Towards Enabling Listening while Speaking in Real-time}
\name{Sashi Novitasari$^1$, Andros Tjandra$^1$, Tomoya Yanagita$^1$, Sakriani Sakti$^{1,2}$, Satoshi Nakamura$^{1,2}$}

\address{
 $^1$Nara Institute of Science and Technology, Japan\\
 $^2$RIKEN, Center for Advanced Intelligence Project (AIP), Japan}
\email{author@university.edu, coauthor@company.com}
\email{\{sashi.novitasari.si3,andros.tjandra.ai6,yanagita.tomoya.yo8,ssakti,s-nakamura\}@is.naist.jp}

\begin{document}

\maketitle
\begin{abstract}
Inspired by a human speech chain mechanism, a machine speech chain framework based on deep learning was recently proposed for the semi-supervised development of automatic speech recognition (ASR) and text-to-speech synthesis (TTS) systems. However, the mechanism to listen while speaking can be done only after receiving entire input sequences. 
Thus, there is a significant delay when encountering long utterances. 
By contrast, humans can listen to what they speak in real-time, and if there is a delay in hearing, they won't be able to continue speaking. 
In this work, we propose an incremental machine speech chain towards enabling machine to listen while speaking in real-time. Specifically, we construct incremental ASR (ISR) and incremental TTS (ITTS) by letting both systems improve together through a short-term loop.
Our experimental results reveal that our proposed framework is able to reduce delays due to long utterances while keeping a comparable performance to the non-incremental basic machine speech chain.
\end{abstract}

\vspace{-0.2cm}
\section{Introduction}
\vspace{-0.15cm}

Automatic speech recognition (ASR) and text-to-speech synthesis (TTS) systems are crucial for human-machine interaction. ASR represents the speech perception system, while TTS represents
the speech production system.
Researchers have been working on ASR and TTS technology for many decades. Current state-of-the-art ASR \cite{bahdanau_2016_1,chiu_2018_1,pham_2019_1} and TTS \cite{wang_2017_1,chen_2019_1} consist of attention-based sequence-to-sequence (seq2seq) neural networks for end-to-end processing.

Despite its remarkable performance, the development of ASR and TTS has progressed more or less independently of each other. Motivated by the human speech chain mechanism \cite{denes_1993_1}, which has a feedback loop phenomenon between speech production and a hearing system, a machine speech chain framework was recently proposed for the simultaneous development of ASR and TTS systems \cite{tjandra_2020_1,tjandra_2017_1,tjandra_2018_1,tjandra_2019_1}. In the machine speech chain framework, both ASR and TTS components are pre-trained in supervised training using a limited amount of labeled data. Then, by establishing a closed feedback loop between the listening component (ASR) and speaking component (TTS), both components can assist each other in unsupervised learning. This loop enables a machine that can learn, not only to listen or speak but also to listen while speaking.

However, standard attention-based ASR and TTS require attention to whole input sequences while producing output. Also, current machine speech chain uses only standard attention-based ASR and TTS components that do not work incrementally. Therefore, the mechanism to listen while speaking can be done only after receiving entire input sequences. ASR starts recognition after receiving a complete speech utterance from TTS, and TTS begins its synthesis after receiving a complete sentence from ASR. As a result, there is a significant delay when encountering long utterances.

By contrast with machines, humans can listen to what they speak in real-time. When perceiving their own speech, speakers simultaneously evaluate their speech and generate their next utterances based on their evaluations of their perceived speech. Several studies have investigated the importance of auditory feedback in speech perception as well as in speech production \cite{parkell_1997_1,badian_1979_1,kurihara_2012_1}. It is done by constructing a delayed auditory feedback (DAF) device that extends the time between speech and auditory perception. A study by Badian et al. \cite{badian_1979_1} found that using DAF with a 175-millisecond delay has been shown to induce mental stress, which was measured as changes in biochemical and cardiovascular variables. Another study also found that the effect of a few hundred milliseconds delay can disturb people, and this effect disappears immediately by stopping speaking \cite{kurihara_2012_1}. Thus, if there is a delay in hearing, humans are unable to continue their speech.

In this work, we attempt to mimic the human speech chain mechanism closely and reduce the delay of feedback within the machine speech chain.
The challenge is that the generation or recognition, and feedback of the spoken utterances must be done based on incomplete sequence information with minimal delay.
Therefore, we propose an incremental machine speech chain mechanism to improve the learning quality of end-to-end incremental ASR (ISR) and incremental TTS (ITTS) through a short-term closed loop.
The proposed mechanism also aims to enable real-time feedback generation during inference.
By enabling the real-time feedback generation, we can move a step closer to achieve an ASR or TTS that is able to adapt simultaneously to the environment unsupervisedly, similar to human.

\vspace{-0.1cm}
\section{Related Works}
\vspace{-0.15cm}

Many researches on ISR and ITTS have been conducted in order to develop a simultaneous speech translation system. The aim is to construct a speech translation system that can mimic human interpreters and translate incoming speech from a source language to the target language in real-time \cite{mieno_2015_1,nakamura_2019_1,fugen_2007_1}.

Several studies on the conventional ASR approach based on the hidden Markov model (HMM) and hybrid systems \cite{saraclar_2002_1,sak_2010_1,lavania_2017_1,peddinti_2018_1} have shown that the framework can recognize speech in real-time. However, the HMM-based ASR cannot perform end-to-end recognition, which is the current state-of-the-art approach with deep learning. By contrast, research on neural ISR is still very limited. Jaitly et al. \cite{jaitly_2016_1} might be the first group that has proposed a neural transducer framework that can recognize speech segment-by-segment with a fixed window. Another study has investigated attention-transfer ISR (AT-ISR) \cite{novitasari_2019_1}, which learns from attention-based non-incremental ASR for end-to-end speech recognition with a low delay.

For TTS, developing ITTS is very challenging; the standard framework commonly requires language-dependent contextual linguistics of a full sentence to produce a natural-sounding speech waveform. Existing studies of ITTS have mainly been conducted on a model based on HMM \cite{baumann_2012_1,baumann_2014_1,pouget_2015_1,yanagita_2018_1}. The first study that attempted to synthesize speech in real-time using neural ITTS was proposed by Yanagita et al. \cite{yanagita_2019_1}. Recently, another ITTS was proposed based on a prefix-to-prefix framework \cite{ma_2019_1}.

Those previously published works were only concerned with ISR and ITTS tasks individually. By contrast, this study investigates the joint of incremental learning between ISR and ITTS that attempts to mimic the process in the human speech chain. The idea here is to construct well performing low-delay systems by evaluating the short-term output of a system using another system and jointly improving both of them. The mechanism of the incremental machine speech chain is based on a basic machine speech chain \cite{tjandra_2017_1} and the incremental steps are learned through attention transfer \cite{novitasari_2019_1}. 
In this work, we perform the inference process with separated ASR and TTS, and also inference with connected ASR and TTS to do a feedback generation.

\vspace{-0.1cm}
\section{Basic Machine Speech Chain}
\vspace{-0.3cm}

\begin{figure}[h]
\vspace{-0.25cm}
 \centering
 \includegraphics[width=0.37\textwidth]{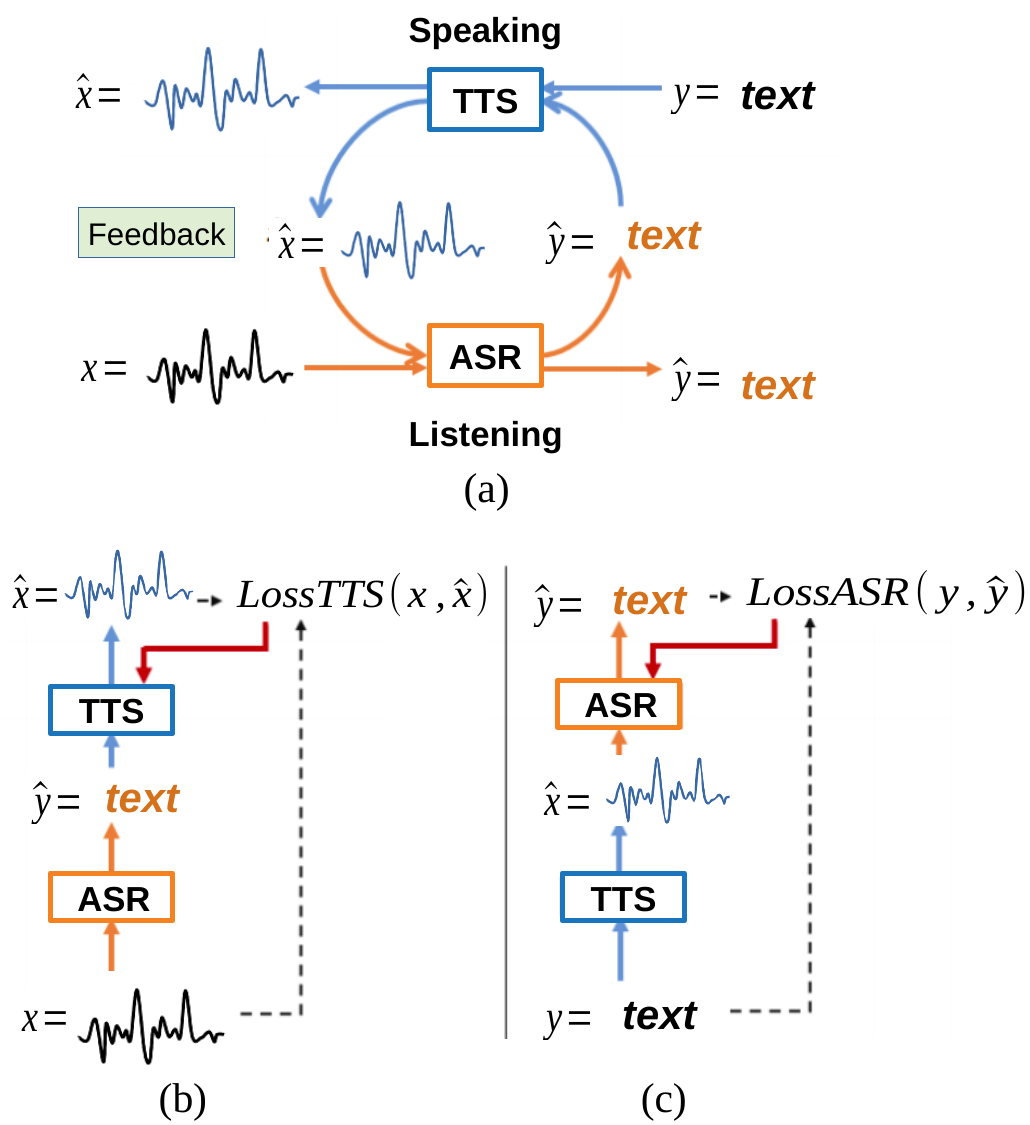}
 \vspace{-0.25cm}
 \caption{Overview of machine speech chain \cite{tjandra_2017_1} (a). The feedback loop is unrolled into two processes: ASR to TTS (b) and TTS to ASR (c).}
 \label{fig:std_msc}
\end{figure}

Fig.~\ref{fig:std_msc} illustrates an overview of the machine speech chain \cite{tjandra_2020_1,tjandra_2017_1,tjandra_2018_1,tjandra_2019_1}.
Machine speech chain trains seq2seq ASR and TTS together by connecting them via a loop. 
Both ASR and TTS frameworks are based on seq2seq neural networks that consist of encoder and decoder components with an attention mechanism \cite{bahdanau_2014_1}.

Machine speech chain was originally proposed for semi-supervised learning for ASR and TTS that consists of two stages: supervised and unsupervised training. 
The supervised training stage is done to train ASR and TTS independently with a small amount of labeled training data. 
The labeled training data consist of pairs of speech and transcribed text. 
This stage acts as a knowledge initialization phase for both components.

After the supervised training phase, ASR and TTS are trained jointly by using the pre-trained models and a large number of unlabeled training data. 
Unlabeled data is data that either speech without transcribed text or text without corresponding speech.
In this phase, ASR and TTS support each other by doing feedback passing through a loop that connects them. The loop between ASR and TTS consists of two unrolled processes:

\vspace{-0.1cm}
\begin{itemize}
 \item \textbf{ASR-to-TTS}.
 ASR transcribes a speech utterance $\mathbf{X}$, with a length $S$, into a sentence text $\mathbf{\hat{Y}}$, and then TTS generates a speech utterance $\mathbf{\hat{X}}$ based on ASR output $\mathbf{\hat{Y}}$. A training loss is calculated based on the original speech $\mathbf{X}$ and TTS speech $\mathbf{\hat{X}}$.
 \item \textbf{TTS-to-ASR}.
 Given a completed sentence text $\mathbf{Y}$ with a length $T$, TTS generates speech $\mathbf{\hat{X}}$ and ASR transcribes the TTS speech $\mathbf{\hat{X}}$ into text $\mathbf{\hat{Y}}$.
 A loss is calculated based on the original text $\mathbf{Y}$ and ASR output text $\mathbf{\hat{Y}}$.
\end{itemize}
\vspace{-0.1cm}

\vspace{-0.15cm}
\section{Incremental Machine Speech Chain}
\vspace{-0.15cm}

An incremental machine speech chain follows the idea of the human speech chain and the basic machine speech chain by establishing a loop that connects ISR and ITTS. The main difference is that the data passing between ISR and ITTS is done within a short time without waiting for a complete utterance.

\vspace{-0.1cm}
\subsection{Components}
\vspace{-0.1cm}
\subsubsection{Seq2seq Incremental Speech Recognition System}
\vspace{-0.1cm}
Seq2seq ISR works with only a low delay through short-segment-based recognition \cite{jaitly_2016_1,novitasari_2019_1}.
ISR predicts sentence text $\mathbf{\hat{Y}}$ with length $T$ from a full speech utterance $\mathbf{X}$ with length $S$ in $N$ recognition steps.
The recognition procedure for each recognition step $n=[1, ..., N]$, where $N=\frac{S}{W}$, is below:

\vspace{-0.05cm}
\begin{enumerate}
 \item Encode $\mathbf{X}_n$, a segment of $W$ speech frames from $\mathbf{X}$, where \($W$ < \)$S$.
 \item Decode and predict $\mathbf{\hat{Y}}_n$, a segment of $K_n$ text tokens from $\mathbf{\hat{Y}}$, where 0 \(\leq\) $K_n$ \(<\) $T$, until an \textit{end-of-block} token predicted by attending encoder states from $\mathbf{X_n}$.
 \item Shift the input window $W$ frames and keep the model states.
\end{enumerate}
\vspace{-0.05cm}

We use attention transfer ISR (AT-ISR) \cite{novitasari_2019_1} in our incremental machine speech chain to limit ISR construction complexity while maintaining reliable recognition performance. 
Attention transfer teaches AT-ISR, the student model, to mimic the alignment from the teacher model or non-incremental ASR that provides $\mathbf{X}_n$-$\mathbf{Y}_n$ pairs based on the attention alignment.
AT-ISR learns $\mathbf{Y}_n$ and an \textit{end-of-block} token as the output target of $\mathbf{X}_n$.
In the attention-based alignment, all $\mathbf{X}_n$ lengths are uniform ($W$), while the length of each text segment $\mathbf{Y}_n$ can be different.

\vspace{-0.1cm}
\subsubsection{Seq2seq Incremental Text-to-Speech Synthesis System}
\vspace{-0.1cm}
Seq2seq ITTS performs speech generation without waiting for a complete sentence text input \cite{yanagita_2019_1,ma_2019_1}.
We construct ITTS using attention transfer from non-incremental ASR.
The non-incremental ASR provides attention-based alignment between fixed-size speech segments and variable-length text segments, which is sufficient to create ISR.
ITTS, with attention transfer from non-incremental ASR, learns how to process a variable-length token sequence to produce at least $W$ speech frames, given the length of a speech segment in the attention-based alignment is $W$.
The ITTS output length here is at least $W$ frames because, during training, we combine subsequent speech segments that did not align to any token ($K_n$=0) to the neighboring segment that aligns with an output token.
We apply the same alignment as ISR for ITTS to reduce components delay incompatibility during joint training.

Speech generation with our ITTS follows the following procedure in each step $n=[1, ..., N]$:

\vspace{-0.05cm}
\begin{enumerate}
 \item Encode $\mathbf{Y}_n$, a segment of $K_n$ tokens from token sequence $\mathbf{Y}$, where 1 \(\leq\) $K_n$ \(<\) $T$.
 \item Decode and predict $\mathbf{\hat{X}}_n$, a segment of $W_n$ speech frames from speech utterance $\mathbf{\hat{X}}$, where $W$ \(\leq\) $W_n$ \(<\) $S$, by attending $\mathbf{Y}_n$ until a stop flag is predicted.
 \item Shift the input window $K_n$ tokens and keep the model states.
\end{enumerate}

\vspace{-0.1cm}
\subsection{ISR and ITTS Training Mechanism}
\vspace{-0.1cm}

\begin{figure}[t]
 \centering
 \includegraphics[width=0.43\textwidth]{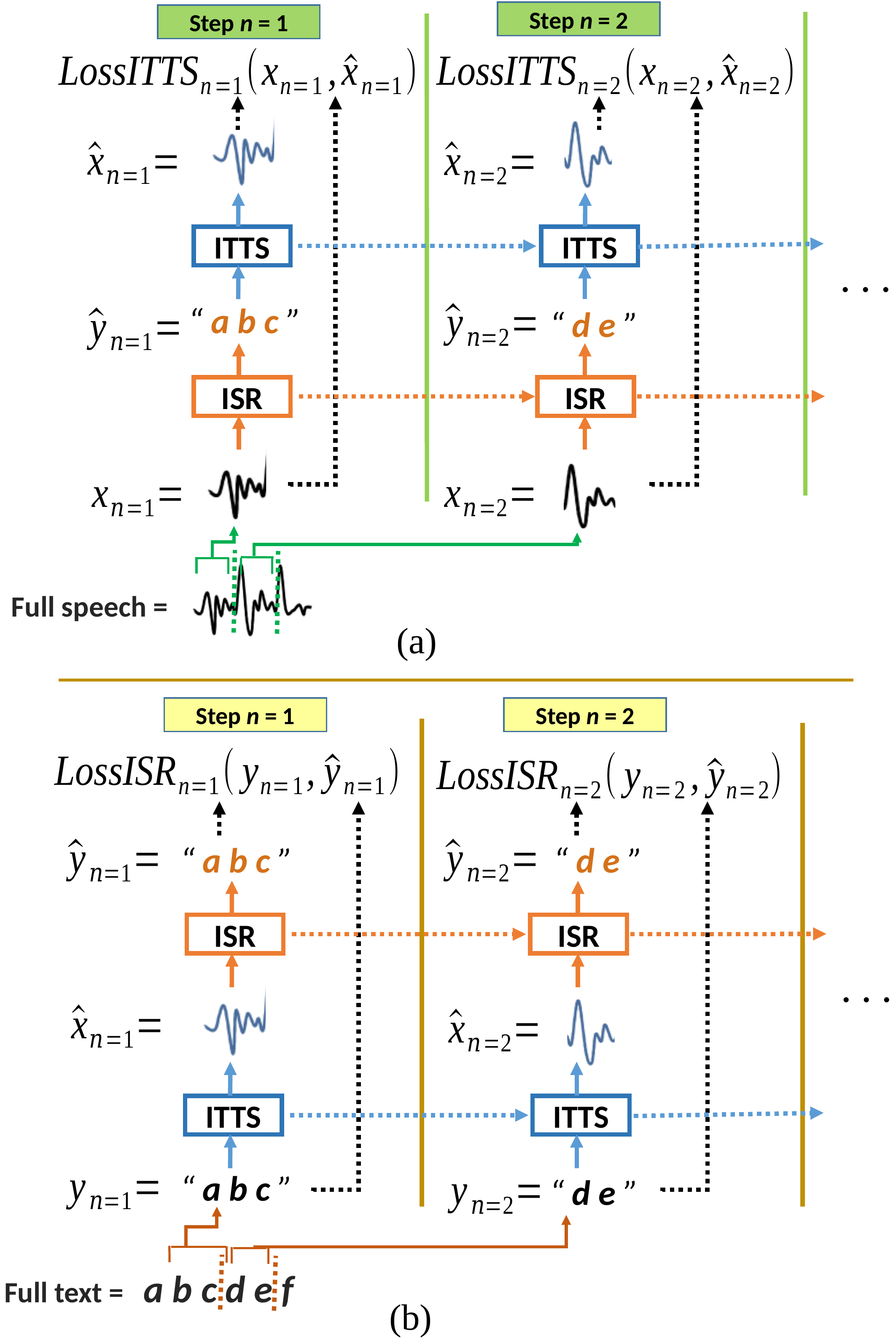}
 \vspace{-0.25cm}
 \caption{Unrolled processes in incremental machine speech chain closed-loop: ISR-to-ITTS (a) and ITTS-to-ISR (b).}
 \vspace{-0.5cm}
 \label{fig:incr_msc}
\end{figure}

Similar to the basic machine speech chain, the training of an incremental machine speech chain also consists of two stages: ISR and ITTS independent training without a closed-loop and ISR-ITTS joint training with a closed-loop.

\vspace{-0.1cm}
\subsubsection{Stage 1: ISR and ITTS Independent-Supervised Training}
\vspace{-0.1cm}
ISR and ITTS are trained independently by using paired speech-text data.
Here attention transfer from non-incremental ASR is applied to train ISR and ITTS, where both of the incremental systems are trained using the same data as the teacher model.

\vspace{-0.1cm}
\subsubsection{Stage 2: ISR-ITTS Joint Training with a Short-term Closed-Loop}
\vspace{-0.1cm}
ISR and ITTS support each other to jointly improve themselves by establishing a short-term closed-loop.
Here each time the first component finishes an incremental step, it passes the output to the second component.
The second component then processes the passed data in an incremental step.
The closed-loop between ISR and ITTS is unrolled into the following processes:

\vspace{-0.05cm}
\begin{itemize}
 \item \textbf{ISR-to-ITTS}.
 In each recognition step $n$, ISR processes a speech segment $\mathbf{X}_n$ and produces an output segment $\mathbf{\hat{Y}}_n$, while ITTS reconstructs the speech segment $\mathbf{\hat{X}}_n$ by encoding $\mathbf{\hat{Y}}_n$, as illustrated in Fig.~\ref{fig:incr_msc}(a).
 Model parameters are updated using averaged ITTS losses from each incremental step, as formulated in Eq.~\ref{eqn:itts_loss}.
\vspace{-0.2cm}
 \begin{equation}
 \text{\textit{LossITTS}} = \frac{1}{N} \sum_{n=1}^{N} \text{\textit{LossITTS}}_n(\mathbf{X}_n,\mathbf{\hat{X}}_n)
 \label{eqn:itts_loss}
 \end{equation}

 \item \textbf{ITTS-to-ISR}.
 For each step $n$, ITTS synthesizes a speech segment $\mathbf{\hat{X}}_n$ by taking a text segment $\mathbf{Y}_n$.
 ISR then transcribes the ITTS speech segment $\mathbf{\hat{X}}_n$ and produces $\mathbf{\hat{Y}}_n$.
 An illustration of this process can be seen in Fig.~\ref{fig:incr_msc}(b).
 Training loss is calculated based on each $\mathbf{Y}_n$ and $\mathbf{\hat{Y}}_n$ pair, which are averaged as can be seen in Eq.~\ref{eqn:isr_loss}.
\vspace{-0.2cm}
 \begin{equation}
 \text{\textit{LossISR}} = \frac{1}{N} \sum_{n=1}^{N} \text{\textit{LossISR}}_n(\mathbf{Y}_n,\mathbf{\hat{Y}}_n)
 \label{eqn:isr_loss}
 \end{equation}
\end{itemize}
\vspace{-0.05cm}

\begin{table*}[h]
\centering
\caption{Performances of ASR (character error rate (CER)) and TTS (L2-norm\textsuperscript{2} between ground truth and predicted Mel spectrogram). 
(nat-sp = natural speech as ASR input; nat-txt = natural text as TTS input;  syn-sp = TTS output as ASR input; rec-txt = ASR output as TTS input; indep-trn: independent training; chain-trn-greedy: joint training via speech chain with greedy intermediate output generation; chain-trn-teachforce: joint training via speech chain with teacher-forcing intermediate output generation).}
\label{tbl:expr_res}
\resizebox{\textwidth}{!}{
\begin{tabular}{|l|cc|cc||cc|cc|}
\hline
\multicolumn{1}{|c|}{} & \multicolumn{4}{c||}{\textbf{ASR (CER(\%))}} & \multicolumn{4}{c|}{\textbf{TTS (L2-norm\textsuperscript{2})}} \\ \cline{2-9} 
\multicolumn{1}{|c|}{} & \multicolumn{2}{c|}{\begin{tabular}[c]{@{}c@{}}\textbf{Non-incremental}\\ (delay: 7.88 sec)\end{tabular}} & \multicolumn{2}{c||}{\begin{tabular}[c]{@{}c@{}}\textbf{Incremental}\\ (delay: 0.84 sec)\end{tabular}} & \multicolumn{2}{c|}{\begin{tabular}[c]{@{}c@{}}\textbf{Non-incremental}\\ (delay: 103 characters)\end{tabular}} & \multicolumn{2}{c|}{\begin{tabular}[c]{@{}c@{}}\textbf{Incremental}\\ (delay: 30 characters)\end{tabular}} \\ \cline{2-9} 
\multicolumn{1}{|c|}{\multirow{-3}{*}{\textbf{Data}}} & {\textbf{nat-sp}} & {\textbf{syn-sp}} &  {\textbf{nat-sp}} & {\textbf{syn-sp}} & {\textbf{nat-txt}} & {\textbf{rec-txt}} & {\textbf{nat-txt}} & {\textbf{rec-txt}}\\ \hline
\multicolumn{9}{|l|}{\textbf{ASR and TTS with independent training}}\\ \hline
{indep-trn (\textit{SI-84})}  & {17.33} &{27.03} & {17.81} & {44.54}  & {0.99} & {1.02} & {1.04} & {3.62}  \\ \hline
{indep-trn (\textit{SI-284})} & {7.16}  &{9.60} & {7.97}  &  {19.99} & {0.75} & {0.77} & {0.84} & {1.31}  \\ \hline
\multicolumn{9}{|l|}{\textbf{ASR and TTS with machine speech chain}}\\ \hline
{indep-trn (\textit{SI-84}) + chain-trn-greedy (\textit{SI-200})} & {11.21} & {11.52} & {14.23} & {32.43} & {0.80} & {0.82}  & {0.86} & {1.35} \\ \hline
{indep-trn (\textit{SI-84}) + chain-trn-teachforce (\textit{SI-200})}      & {7.27}  & {6.30} & {9.43} & {12.78}  & {0.77}  &  {0.80} & {0.79} & {1.26}  \\ \hline
\end{tabular}
}
\end{table*}

\vspace{-0.1cm}
\subsection{Incremental Machine Speech Chain Learning Approach}
\vspace{-0.1cm}
Basic machine speech chain framework was originally proposed for semi-supervised learning, in which the intermediate output in an unrolled feedback loop process is generated through a greedy decoding mechanism.
In this work, as our focus is not on semi-supervised learning, we explore two approaches for intermediate output generation during joint training via closed-loop: teacher-forcing approach and greedy approach.

In the joint training that synthesizes the intermediate output with a teacher-forcing approach, the first system in an unrolled-loop process generates an output sequence through a teacher-forcing decoding mechanism.
The joint training here is done using labeled data.

Joint training that generates the intermediate output with a greedy approach is done using unlabeled data. 
Here, the output generation in the first system of an unrolled-loop process is done through a greedy decoding mechanism.

\vspace{-0.10cm}
\section{Experiment Setting}
\vspace{-0.20cm}

\subsection{Dataset}
\vspace{-0.1cm}
We used Wall Street Journal (WSJ) \cite{paul_1992_1} dataset for ASR and TTS construction with the following setting: \textit{SI-84}, \textit{SI-200}, and \textit{SI-284} as the training sets, \textit{dev93} as the development set, and \textit{eval92} as the test set.
The \textit{SI-84} set consisted of 16 hours of speech by 83 speakers and the \textit{SI-200} set consisted of 66 hours of speech by 200 speakers that did not overlap with \textit{SI84} set.
The \textit{SI-284} set was a combination of \textit{SI-84} and \textit{SI-200} sets.
The \textit{SI-84} and \textit{SI-284} were utilized to train the ASR and TTS during independent training, while the \textit{SI-200} set was utilized for systems joint training with a closed-loop.   
All speech utterances had a sampling rate of 16-kHz.

For both ASR input and TTS output,
speech utterances were represented as 80-dimension log Mel spectrograms, where each feature frame had a length of 50 ms that shifted by 12.5 ms from the previous frame.
In the ASR output and TTS input sides, the text data was represented as a sequence of character units.

\vspace{-0.10cm}
\subsection{Model Configuration}

\vspace{-0.1cm}
\subsubsection{ASR}
\vspace{-0.1cm}
Non-incremental ASR and ISR had identical seq2seq structures.
The encoder consisted of a feed-forward neural network layer (512 units) that was followed by three bidirectional LSTM layers (256 units each).
Each bidirectional LSTM layer applied hierarchical sub-sampling \cite{bahdanau_2016_1,graves_2012_1}.
As a result, an encoder state in the encoder's final layer represented eight speech frames.
Here we defined eight speech frames (0.14 sec) as a speech block.

The ASR decoder consisted of a character embedding layer (256-dims), an LSTM layer (512 units) with an attention mechanism, and a softmax layer.
We applied an MLP-scoring function that used a previously proposed multi-scale alignment and contextual history \cite{tjandra_2018_2} in the attention component.
The text generation during inference was done by greedy-decoding to prevent an additional delay.

\vspace{-0.2cm}
\subsubsection{TTS}
\vspace{-0.1cm}
Our TTS followed the TTS structure in previous machine speech chain work \cite{tjandra_2017_1}, which was a modification from TTS Tacotron \cite{wang_2017_1}.
The model hyperparameters were generally same as those in the original Tacotron.
The modification was made by replacing the rectified linear unit (ReLU) function with the leaky ReLU (LReLU) function.
The CBHG (1-D convolution bank + highway network + bidirectional GRU) module used $K = 8$ filter banks.
The decoder consisted of two LSTM (256 units) layers.
Our TTS generated 4 consecutive frames for each decoding step, thus reducing the number of total decoding steps.

Our TTS implemented a speaker recognition component similar to that in previous machine speech chain work \cite{tjandra_2018_1}.
Here we used a DeepSpeaker model \cite{li_2017_1} with the same hyperparameters as in the previous machine speech chain research.
The speaker recognition component was trained using \textit{SI-84} set.

\vspace{-0.10cm}
\section{Experimental Results and Analysis}
\vspace{-0.10cm}
Our experiment result can be seen in Table~\ref{tbl:expr_res}.
The baselines are ISR and ITTS that were trained independently using \textit{SI-84} set.
The toplines are the non-incremental systems that were trained independently with \textit{SI-284} set.
The machine speech chain mechanism for the non-incremental systems followed the basic mechanism (see Sec. 3), while the incremental systems followed the incremental mechanism (see Sec. 4).
Here the system evaluation was done based on natural input and synthetic input.
The synthetic input was generated by the target system's counterpart system, which was trained under the same training condition, by processing a natural input.
For the incremental system, the synthetic input of an incremental step was the output from the processing of a short segment of natural input.
We can consider the output of synthetic input processing as the feedback for the system that produced the synthetic input.

The results show that our proposed framework could reduce the delay when encountering long utterances with a close performance to the non-incremental speech chain.
Here we allowed the ISR and ITTS to take contextual inputs, which consisted of look-back and look-ahead blocks \cite{novitasari_2019_1,yanagita_2019_1,ma_2019_1,sainath_2018_1} that are the blocks before and after the main segment respectively, to enrich the information in the main input segment.
The ISR input segment for an incremental step consisted of four main speech blocks with two look-back and four look-ahead blocks, which we decided based on optimum delay configuration in previous AT-ISR work \cite{novitasari_2019_1}. The ISR delay here was equal to 0.84 sec.
The ITTS input segment in an incremental step consisted of the main character block with two look-back and four look-ahead character blocks.
The ITTS main input size range was between one to four blocks, with an average of two blocks.
Here a character block consisted of five characters, the average word length in the training data.
Our best proposed incremental systems were ISR and ITTS that were trained jointly by applying teacher-forcing approach to generate the intermediate output.
Given a natural input, the best proposed ISR achieved 9.43\% CER with a delay of 0.84 sec, while the non-incremental ASR that was trained with the basic machine speech chain achieved a CER of 7.27\% but with a delay of 7.88 sec.
The best ITTS, given a natural input, was able to predict speech features with a loss of 0.79 to the ground truth by only waiting for 30 characters on average, while the non-incremental TTS with the same training data achieved a synthesis loss of 0.77 but has to wait for 103 characters on average to begin the synthesis.

Incremental machine speech chain framework successfully improves ISR and ITTS performances.
The improvement of ISR and ITTS occurred on both natural input and synthetic input processing.
It shows that the short-term feedback loop between the incremental systems are able to leverage their training quality.
Here we also demonstrated a real-time feedback generation.
This is an important step to achieve a system that can listen while speaking in real-time.

\vspace{-0.25cm}
\section{Conclusion}
\vspace{-0.20cm}
Our aim was to mimic the process of the human speech chain by constructing an incremental machine chain and allowing ISR and ITTS to improve together through a short-term loop. 
Our experimental results reveal that our proposed framework can reduce the delay when encountering long utterances while keeping a close performance to a non-incremental speech chain and also outperforming the baseline ISR and ITTS.

\vspace{-0.20cm}
\section{Acknowledgement}
\vspace{-0.20cm}
Part of this work is supported by JSPS KAKENHI Grant Numbers JP17H06101 and JP17K00237.

\bibliographystyle{IEEEtran}

\bibliography{mybib}


\end{document}